\documentclass[a4paper,10pt]{esannV2}
\usepackage[a4paper,left=44mm,right=44mm,top=53mm,bottom=54mm,
            headheight=0pt,headsep=0pt]{geometry}
\usepackage{amsmath,amssymb,graphicx,booktabs}
\usepackage[protrusion=false]{microtype}
\usepackage[hidelinks]{hyperref}
\newcommand{\relu}{\operatorname{ReLU}}
\newcommand{\softplus}{\operatorname{softplus}}

\title{Compressing probabilistic almost-linear recurrent networks without refitting}
\author{F. Olivia Fan, Oliver Obst\vspace{.3cm}\\
School of Computer Science and Engineering\\
UNSW Sydney, Australia}

\begin{document}
\maketitle

\begin{abstract}
Can a compressed recurrent network reuse its learned noise model? We applied finite-horizon balanced truncation to ten probabilistic almost-linear networks on delayed recall. Reducing 64 states to 32 without refitting met forecast tolerances for seven networks when balancing included rectifier inputs, compared with one using the readout alone. Including rectifier inputs improved agreement with the parent's mean-centred forecast distribution on all ten networks, with a median reduction in discrepancy of 65\%. Compared with random directions, rectifier inputs reduced centred divergence on six networks and mean forecast error on all ten. Both full and reduced networks underestimated uncertainty on weak cues.
\end{abstract}

\section{Introduction}
If two states with different future output distributions project to the same point, a forecaster given only that point cannot recover both distributions. Which state directions let a smaller network reuse its learned noise model? Stochastic model reduction distinguishes several targets, including matching response means and higher moments~\cite{stochasticmoments}. Here the target is the distribution of a short output trajectory conditional on an input sequence. It matters, for instance, whether a recalled signal keeps the same sign over four consecutive outputs.

In an almost-linear recurrent network (AL-RNN), a small set of rectifiers supplies the nonlinear part of the recurrence~\cite{alrnn}. Earlier piecewise-linear RNNs placed noise on the recurrent state~\cite{plrnn}. We place it at the rectifier inputs and learn its scale from data. A state direction with little immediate effect on the readout can change a rectifier input and affect the distribution of later outputs.

Besselink et al.~\cite{lure} reduce deterministic feedback systems by balancing the linear dynamics with both the readout and the inputs to the nonlinearity as observations, then reconnecting the nonlinearity. We use the finite-horizon version of linear balancing~\cite{timebalance} to study a different target: the conditional distribution produced by a trained stochastic recurrence. We compare the readout alone, the readout with rectifier inputs, and the readout with the same number of random state directions. A known posterior lets us distinguish agreement with the parent from accuracy on the task.

\section{The stochastic recurrence and its reduction}
Let $z_t\in\mathbb R^N$ be the state, $u_t$ the external input, and $E$ the map from the state to $q$ rectifier inputs. The full network selects its last $q$ coordinates with $E$. Starting from zero, we use
\begin{align}
s_t &= \softplus(Sz_t+b), \qquad \varepsilon_t\overset{\rm iid}{\sim}\mathcal N(0,I_q), \nonumber\\
\phi_t &= \relu(Ez_t+s_t\odot\varepsilon_t), \nonumber\\
z_{t+1} &= Az_t+W\phi_t+Bu_{t+1}+h, \qquad y_t=Dz_t . \label{eq:full}
\end{align}
Here $A$ is a trainable dense matrix. Conditional on $z_t$, gate $j$ opens with probability $\Phi((Ez_t)_j/s_{t,j})$. Noise can change both the variance and the mean of the forecast, even when $A,W,B,h,D$ are frozen.

\paragraph{Transporting a fitted network.}
Choose maps $Q\in\mathbb R^{M\times N}$ and $V\in\mathbb R^{N\times M}$ with $QV=I_M$. The $M$-state network uses the same recurrence with
\begin{equation}
\begin{split}
A_r=QAV,\quad W_r=QW,\quad B_r=QB,\quad h_r=Qh,\\
E_r=EV,\quad S_r=SV,\quad D_r=DV,\quad b_r=b .
\end{split}\label{eq:transport}
\end{equation}
It stores the reduced matrices and evaluates the gates and their noise scales directly from its own state. All $q$ gates remain, including when $M<q$.

\paragraph{An exact case.}
Suppose the rows retained by $Q$ satisfy
\begin{equation}
QA=A_rQ,\qquad E=E_rQ,\qquad S=S_rQ,\qquad D=D_rQ .\label{eq:exact}
\end{equation}
For any given input, initialise $r_0=Qz_0$ and drive both networks with the same innovations. If $r_t=Qz_t$, then the gate inputs and noise scales agree, so both networks produce the same $\phi_t$. Applying $Q$ to~\eqref{eq:full} gives $r_{t+1}=Qz_{t+1}$. Induction and $D=D_rQ$ give equal outputs on every path, hence equal joint output laws. The discarded coordinates can be driven by the input and vary from step to step. In a constructed eight-state example satisfying~\eqref{eq:exact}, four retained states reproduced 1,024 sampled paths of length 41 with a maximum absolute output difference below $1.2\times10^{-15}$.

\paragraph{Choosing directions in a trained network.}
We choose balanced directions of the linear recurrence over the 41-step task horizon. The external inputs, gate outputs and bias drive the linear state through $F=[B\ W\ h]$. The observation matrix $L$ specifies which state measurements should be preserved. With $H=41$, the reachability and observability factors are
\begin{equation}
\mathcal R=[F,AF,\ldots,A^{H-1}F],\quad
\mathcal O=[L^\top,(LA)^\top,\ldots,(LA^{H-1})^\top]^\top .
\end{equation}
The leading $M$ singular directions of $\mathcal O\mathcal R$ give the two balanced projection maps~\cite{timebalance}. Treating the gate outputs as external drives makes the choice of basis tractable. We test its effect in the complete stochastic recurrence.
\section{Delayed recall with a known posterior}
Each episode encodes a signed memory $m=\eta a$, where $\eta$ is a fair sign and $a$ is uniform on $[0.25,1]$. Eight observations are $\eta(r_j-\bar r)+m/8+e_j$, with independent $r_j\sim\mathcal N(0,0.1^2)$ and $e_j\sim\mathcal N(0,0.01^2)$. After a delay chosen uniformly from 12 to 28 steps, a timing pulse marks a cue for a second independent fair sign $c$. We use a noiseless cue $c$ on 70\% of episodes. On the remaining episodes the cue is $c\kappa+\xi$, with $\kappa$ uniform on $[0,0.3]$ and $\xi\sim\mathcal N(0,0.1^2)$. The four targets, one through four steps after the cue, are $cm(1,0.75,0.5,0.25)$.

The encoding sum equals $m$ plus Gaussian noise of variance $8(0.01)^2$. The other seven contrasts are independent of $m$, so its posterior is a Gaussian centred at the observed sum, restricted to the two prior intervals. The noiseless cue identifies $c$. For a weak cue we integrate its Gaussian likelihood over $\kappa$ for each sign. A cue near zero can support either sign, while a larger weak cue can favour one. Each oracle draw uses the same posterior draw of $(c,m)$ for all four outputs. No hidden sign or cue label enters the forecaster or the oracle.

\paragraph{Training and evaluation.}
We use 64 states and 16 gates. We fit $A,W,B,h,D$ for 400 Adam updates on squared error with innovations set to zero. We then freeze them and fit $S,b$ for 200 updates on the energy score of four sampled output trajectories~\cite{gneiting}. Each update uses 32 episodes, learning rate 0.003 and unit gradient clipping. Each full network trains on 1,500 episodes. The ten-network comparison uses independent training and test sets for each initialisation. Checkpoints and ranks are fixed before test episodes are generated. On every test input we sample 128 trajectories per model, with common innovations for full and projected networks.

We report the RMSE of the sampled predictive mean, marginal continuous ranked probability score (CRPS), coverage of central marginal 90\% intervals, and joint energy score (ES) on the four-dimensional target. Lower scores are better. For both proper scores we exclude identical sample indices from estimates of within-ensemble distances. A reduced model passes on a seed if coverage is in $[0.85,0.95]$, RMSE is at most $0.25$ and at most $0.025$ above its parent, CRPS is at most $0.015$ above its parent, and ES at most $0.04$ above its parent. We report how many networks meet all four tolerances, which we fixed before testing.

\section{Which state directions matter at half the dimension?}
\paragraph{Choice of rank.}
Two preliminary three-network studies motivated a comparison at 32 states. In the first, balanced reduction to 60 states met the forecast tolerances on all three networks without refitting. At 13 states, even the primary procedure, which selected spectral modes by omission error~\cite{lrnn} and refitted dynamics and noise scale, passed on just one. The second study tested intermediate ranks on fresh networks. At 40 states, balancing with $D,E$ passed on all three: mean CRPS was 0.063 against 0.060 at full size. At 32 states, using $D,E$ passed on two and additionally observing $S$ (the primary comparison) passed on one. This left open whether observing the rectifiers helped chiefly by preserving the mean, and whether their particular directions mattered.

\paragraph{Paired experiment.}
We fixed the following comparison before training ten new 64-state parents. Each parent supplies five 32-state reductions: one observes the readout $D$, one observes $D$ and the 16 rectifier inputs $E$, and three observe $D$ and 16 random orthonormal state directions. The random rows have the same norms as the rows of $E$ and are used only to select the basis. All five reductions use the same driving matrix $F$, the actual gate and scale maps transformed by~\eqref{eq:transport}, and zero optimisation steps. We froze all 60 checkpoints before generating 600 held-out episodes per parent. The unit of comparison is a trained network with its own data. We average the three random controls' scores within each network.

\paragraph{Separating the mean from the rest of the forecast.}
For each test input, we estimate each model's four-dimensional forecast mean on 64 draws and subtract it from the other 64 draws. We compare the two centred ensembles using the energy divergence~\cite{szekely}, swap the halves and average. The divergence is twice the mean distance between ensembles less the two mean distances within ensembles. We exclude equal draw indices because the models share innovations. Two further measures cancel additive shifts exactly: the error in central interval width and the divergence between distributions of differences of independent trajectories, scaled by $1/\sqrt{2}$.

We prespecified a material improvement as lower centred divergence on at least nine of ten networks, with median paired reductions of at least 0.0005 in absolute terms and 20\% in relative terms. Both shift-invariant measures also had to improve in their arm medians and paired median differences. We applied the same rule against the random control. Separately, to claim comparable unadjusted means we required a paired 90\% interval for the difference in target RMSE within $\pm0.01$, and both reductions' means within 0.05 RMSE of the parent in at least nine networks.

\begin{table}[t]\centering\small
\begin{tabular}{lrrrrr}\toprule
Basis observations & RMSE & CRPS & Cov.\ (\%) & Joint ES & Pass\\\midrule
Full (64 states)      & .175 & .075 & 88.9 & .171 & 9/10\\
Readout               & .258 & .148 & 67.6 & .337 & 1/10\\
Readout and random rows & .222 & .119 & 72.5 & .275 & 0/30\\
Readout and rectifiers & .183 & .082 & 87.5 & .189 & 7/10\\\bottomrule
\end{tabular}
\caption{Forecasts at 32 states without refitting. Scores average ten parent networks with 600 test episodes each. The random row averages three projections per network. Its pass count refers to the 30 individual projections.}\label{tab:main}
\end{table}

\paragraph{Forecast accuracy.}
The reduction using the rectifiers met all four task tolerances on seven networks, compared with one for readout alone (Table~\ref{tab:main}). It gave lower target RMSE than either readout alone or the average random control on every network. The paired difference from readout alone averaged $-0.075$, with a 90\% interval of $[-0.115,-0.036]$. None of the ten pairs had both reduced means within 0.05 RMSE of their parent. Thus the methods differed materially in their unadjusted mean forecasts.

\begin{figure}[t]\centering
\includegraphics[width=\linewidth]{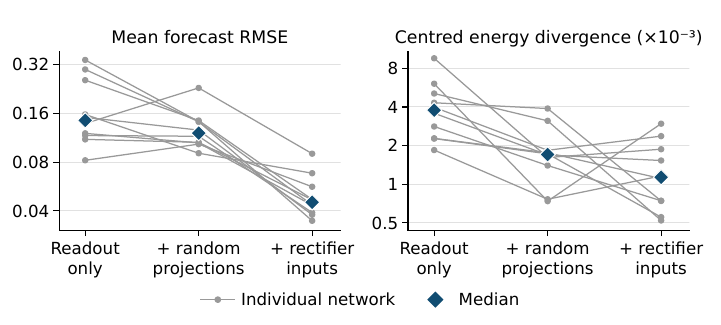}
\caption{Agreement with the full network at 32 states. Each grey line connects reductions of the same network. Diamonds are medians. The left panel measures per-step error in the forecast mean. The right compares the joint forecasts after removing their estimated conditional means. Both axes are logarithmic.}\label{fig:paired}
\end{figure}

\paragraph{Forecast variability.}
After centring, the rectifier-aware reductions came closer to the full model than readout alone on every network (Fig.~\ref{fig:paired}). The median paired decrease in divergence was 0.00206, or 65.4\% in relative terms. Interval-width errors and the divergence for trajectory differences also fell on all ten. Against the random control the rectifiers won only six comparisons. The median absolute gain was 0.00040 and the median relative gain was 23.2\%. Neither shift-invariant measure had a positive median paired gain. Mean interval-width errors relative to the parent were 0.072 for readout alone, 0.043 for random directions and 0.042 for rectifier inputs.

\section{Discussion}
The learned noise map can be carried through a reduction to half the state dimension. Accounting for the rectifier inputs when choosing the subspace gave the most accurate mean forecasts and met the task tolerances on seven of ten networks. Observing additional internal directions also improved agreement in forecast variability. The random controls account for much of this improvement, leaving the role of the specific rectifier directions in preserving variability unresolved.

Across the ten new test sets, 1,816 of 6,000 episodes had weak cues. Block-average coverage of nominal 90\% intervals on weak cues was 63.4\% at full size and 64.4\% after rectifier-aware reduction. On strong cues it was 99.95\% and 97.7\%, respectively. On weak cues, halving the state retained the full network's tendency to underestimate uncertainty.

\end{document}